\documentclass[conference]{IEEEtran}
\usepackage{graphicx}
\usepackage{epstopdf}
\usepackage{array}

\usepackage{placeins}
\usepackage{multirow}
\usepackage{mathtools}
\usepackage{amsmath}
\usepackage{amssymb}
\usepackage{color}
\usepackage{caption}

\usepackage{gensymb}

\usepackage{caption}
\usepackage{subcaption}

\begin{document}
%
\title{A Product {Shape} Congruity Measure via Entropy in Shape Scale Space}

\author{\IEEEauthorblockN{Asli Genctav, Sibel Tari}
\IEEEauthorblockA{Middle East Technical University, Department of Computer Engineering, Ankara, Turkey\\
asli@ceng.metu.edu.tr, stari@metu.edu.tr}}


%


\maketitle

\begin{abstract}
Product shape is one of the factors that trigger preference decisions of customers. Congruity of shape elements and deformation of shape from the prototype are two factors that are found to influence aesthetic response, hence preference. We propose a measure to indirectly quantify congruity of different parts of the shape and the degree to which the parts deviate from a sphere, i.e. our choice of the prototype, without explicitly defining parts and their relations. The basic signals and systems concept that we use is the entropy. Our measure attains its lowest value for a volume enclosed by a sphere. On one hand, deformations from the prototype cause an increase in the measure. On the other hand, as deformations create congruent parts, our measure decreases due to the attained harmony. Our preliminary experimental results are consistent with our expectations.
\end{abstract}


%
\IEEEpeerreviewmaketitle

\section{Introduction}

Product shape is one of the factors that trigger preference decisions.
Marketing research provides evidence that size, proportion, shape affect customer preference. There is also research on the effect of other design principles on preference.
For example, Veryzer and Hutchinson~\cite{veryzer98} investigated prototypicality and unity as factors that influence aesthetic response, hence preference.  We found Veryzer and Hutchinson experiments particularly interesting because their definition of unity as a relational property refers to congruity among the elements of a design whereas prototypicality as an atomic property refers to a deformation.

In our paper, using fundamental concepts from signals and systems we suggest a measure to indirectly quantify congruity of different parts of the shape and the degree to which the parts deviate from simplicity, i.e., mixture of unity and prototypicality, without explicitly defining parts and their relations. In general, product preference assessment experiments are difficult because it is hard to isolate factors and manipulate them orthogonally. Hence, we believe that our indirect implicit measure may be useful in guiding experiments.

The basic signals and systems concept we use is entropy \cite{shannon48} {which is frequently used for measuring complexity. For example, Batty et al. \cite{Batty2014} used entropy for defining a measure of complexity for spatial systems.} {In his book \cite{Birkhoff}, Birkhoff defined aesthetic measure as the ratio between order and complexity of an object, which is closely connected with the aesthetic demand as unity in variety.}

The proposed measure is expected to attain its lowest value for a volume enclosed by a sphere. This is our choice of a prototype.  On one hand, deformations from the prototype cause an increase in the measure. On the other hand, as deformations create parts that are congruent, this causes a decrease due to attained harmony.

Our focus is on 3D volumes as they model solid products. Nevertheless, constructions are valid for shapes in arbitrary dimensions.

\section{Method}

For a random variable $F$ which takes values from a set $E$, the entropy is defined as 
\begin{equation}
- \sum_{f \in E} p_F(f) \, \log_2 p_F(f)
\label{eq:entropy}
\end{equation}
where $p_F$ is the probability distribution of the random variable $F$.
The more uniform is the probability distribution $p_F$, the higher the entropy.

{Assuming product shape is represented in the form of an open subset $V \subset \mathbb{R}^3$ with a bounding surface $S$, we quantify its congruity,  by 
estimating a parameterized collection of entropy values $\hat{e}_{i,k}$ corresponding to sample  realizations $\hat{F}_{i,k}$ of a family of random variables.}
{Here,  $\hat{F}_{i,k}$ are obtained by multiple systematic sampling of 
the values of a one parameter family of real valued fields $v_{\rho_i}: V \rightarrow \mathbb{R}$.  In practice, $V$ can be a point cloud.}
{Each $v_{\rho_i}$ is sampled by collecting values on a particular loci $B_{\delta_k}$ defined as the points in the volume which have a constant predetermined distance $\delta_k$ to the bounding surface $S$. That is $B_{\delta_k} = \{x \mid d(x,S) = \delta_k\}$. For $\delta_k=0$, $B_0$ coincides with $S$.} $\hat{F}_{i,k}$ is
\[  \{v_{\rho_i}(x) \mid x \in B_{\delta_k}  \} \]
{where the corresponding probability distribution is estimated using histogram of sample values of $\hat{F}_{i,k}$.}

{In the proposed scheme, the design of $v_{\rho_i}$ is important. For our purpose,  we employ a previous shape modeling tool \cite{tsp96}. It is designed in a way that for a sphere, equidistant loci $B_{\delta_k}$ coincide with the iso-height surfaces of $v_{\rho_i}$. This means that the elements of $\hat{F}_{i,k}$ are identical, hence, the corresponding entropy is 0.  For an arbitrary shape, the iso-height surfaces of $v_{\rho_i}$ emulates gradual smoothing of $B_0$, {\sl i.e.} the enclosing surface $S$. The parameter $\rho_i$ controls the rate of smoothing. 
The more shape deviates from a prototype (implicitly assumed as the sphere), the more the elements of $\hat{F}_{i,k}$ vary, causing an increase in the entropy.} 
When the deviations yield congruent shape parts, {where we relate congruity to similarity of shape parts in terms of their geometry and configuration}, the sample $v_{\rho_i}$ values collected from those parts contribute similarly to the distribution of $\hat{F}_{i,k}$ {since the shape modeling tool that we use is able to characterize each shape point with respect to local and global shape configuration}.
Thus, we expect the entropy values $\hat{e}_{i,k}$ to be smaller for a shape with congruent parts than a shape with incompatible parts.

\paragraph*{{Constructing $v_{\rho_i}$}}

We obtain $v_{\rho_i}$ by solving the following equation inside the volume enclosed by the surface $S$.
 \begin{equation}
\Delta \, v_{\rho_i} - \frac{1}{{\rho_i}^2} \, v_{\rho_i}  = -1 \mbox{ subject to } v_{\rho_i} \left|_{S} = 0 \right.
 \label{eq:pde}
\end{equation}
Here, $\Delta$ is the Laplace operator and $\rho_i$ is a parameter which controls the smoothness level of $v_{\rho_i}$.

We obtain the solution by discretizing it on a standard grid via finite-difference method. The discretization yields a linear system of equations. We solve the resulting linear system using a direct solver based on Cholesky factorization.

$v_{\rho_i}$ is $0$ on the boundary and increases towards the volume center. We normalize $v_{\rho_i}$ by dividing it to its maximum value, hence, its range is $[0,1]$. Notice that as $\rho_i$ increases, the relative importance of the smoothness term gets higher, and hence, $v_{\rho_i}$ becomes smoother.

\section{Experimental Results and Discussion}

First, we experiment with a set of shapes consisting of a cube and seven composite cubes which are generated as combinations of a base cube and smaller identical cubes. Second, we consider a set of shapes consisting of a sphere and three cubic shapes with differing parts.

\begin{figure}[t]
\centering
\includegraphics[height=0.48\linewidth, trim=180 20 160 20, clip]{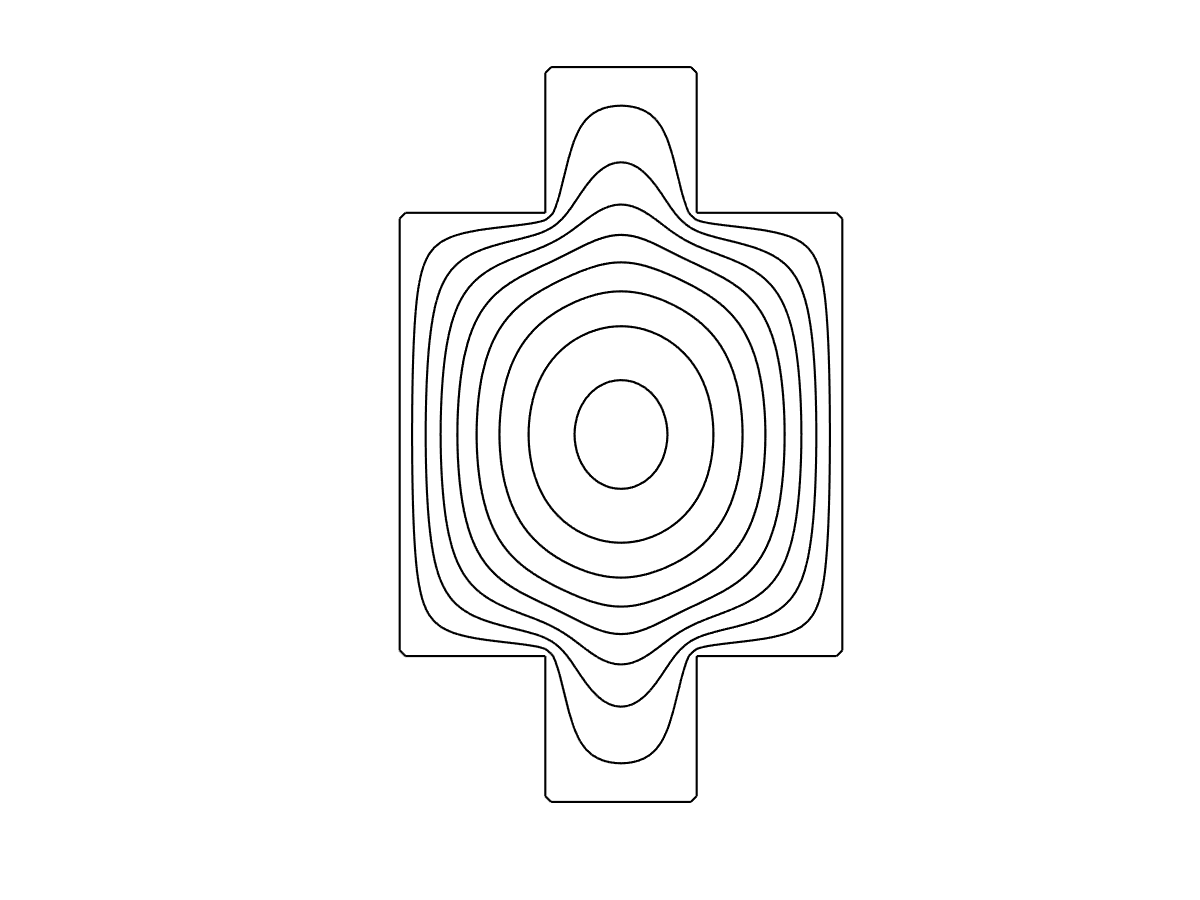}
\includegraphics[height=0.48\linewidth, trim=180 20 90 20, clip]{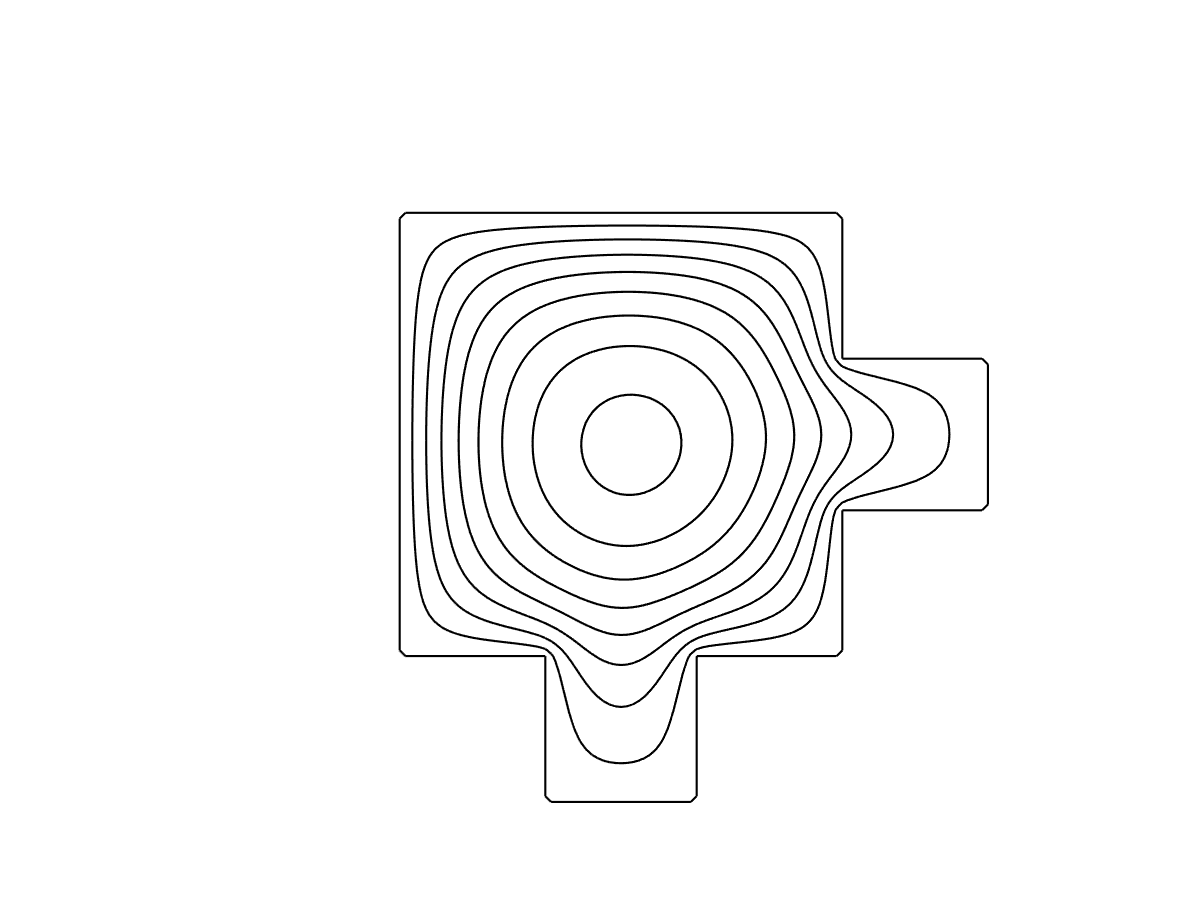}\\
\includegraphics[height=0.48\linewidth, trim=180 20 160 20, clip]{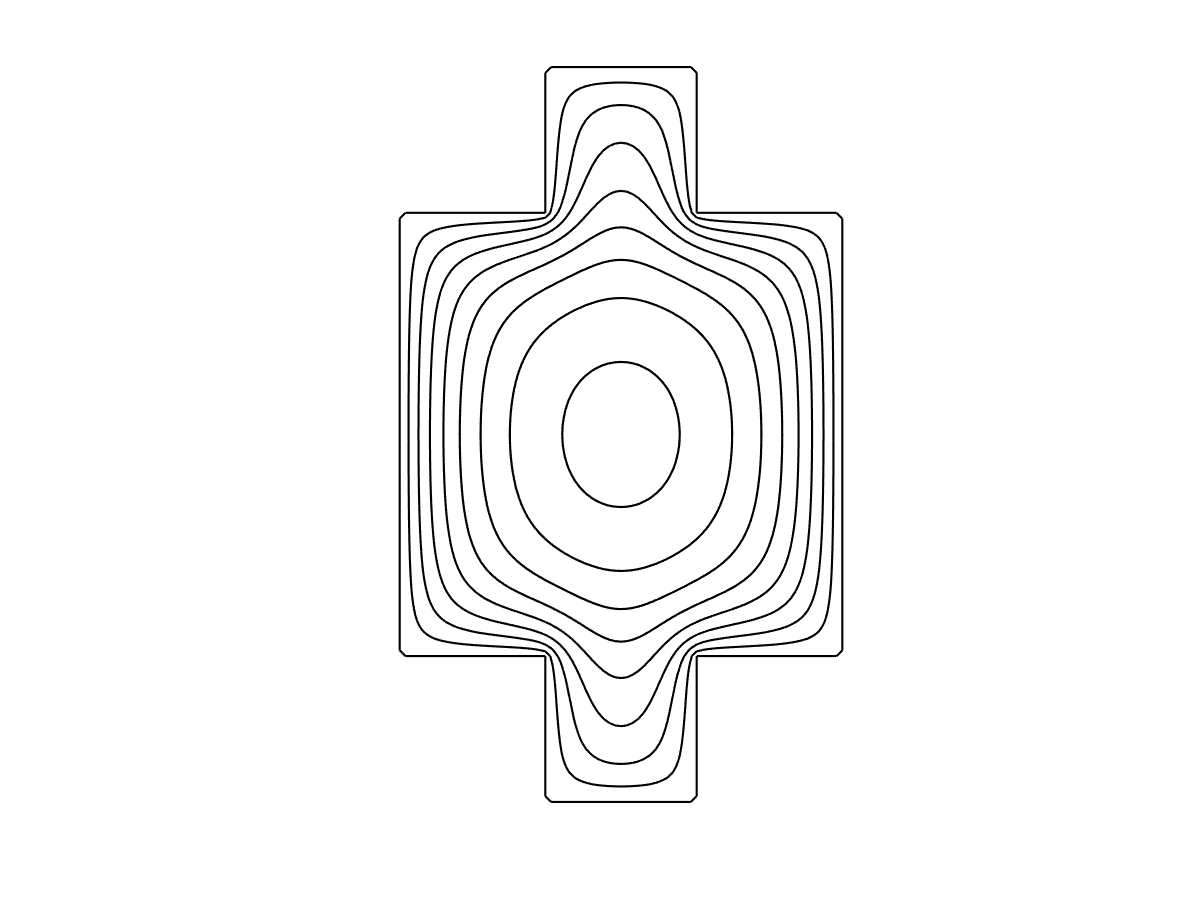}
\includegraphics[height=0.48\linewidth, trim=180 20 90 20, clip]{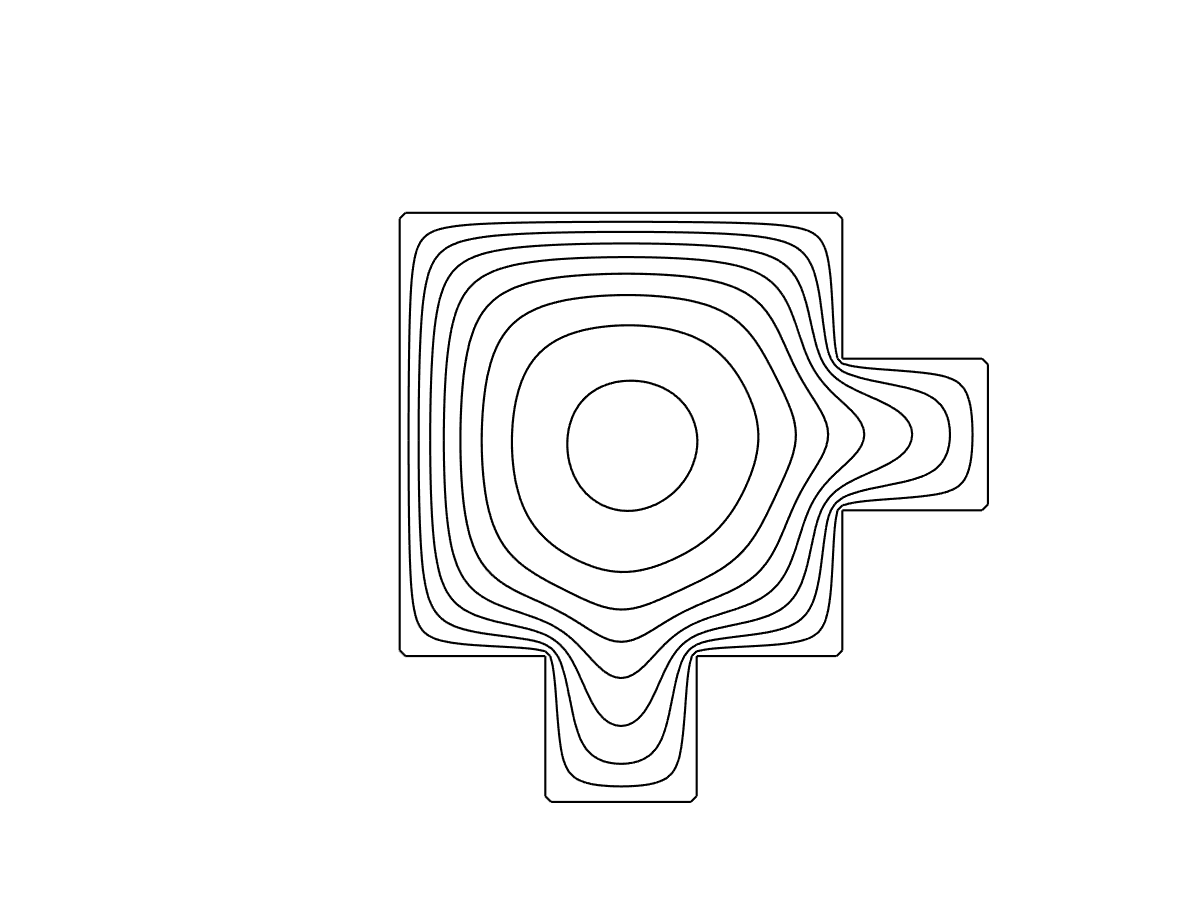}
\caption{{Iso-level curves of $v_{\rho_1}$ (top) and $v_{\rho_2}$ (bottom) for two different 2D shapes. $\rho_1 > \rho_2$ as $\rho_1$ is equal to shape area and $\rho_2$ is equal to ${\rho_1}^{0.3}$.}}
\label{fig_vfield} 
\end{figure}

Using two different selections for $\rho_i$ and $\delta_k$, we obtain four entropy measures that quantify shape congruity of the volumes.
In order to exemplify the shape scale space, $\rho_1$ is chosen as a large value equal to the volume size whereas $\rho_2$ is chosen as a smaller value equal to ${\rho_1}^{0.3}$ which corresponds to approximation of the radius in the case that the volume is a sphere. 
{In Fig.~\ref{fig_vfield}, we show iso-level curves of $v_{\rho_1}$ (top) and $v_{\rho_2}$ (bottom) for two different 2D shapes. Notice that $v_{\rho_1}$ is smoother than $v_{\rho_2}$ as $\rho_1 > \rho_2$.}
We determine $\delta_1$ and $\delta_2$ as $0.05$ and $0.1$ times the maximum thickness of the volume.

\begin{figure*}[!t]
\centering
\begin{subfigure}[b]{0.7\linewidth}
\centering
\includegraphics[width=\linewidth]{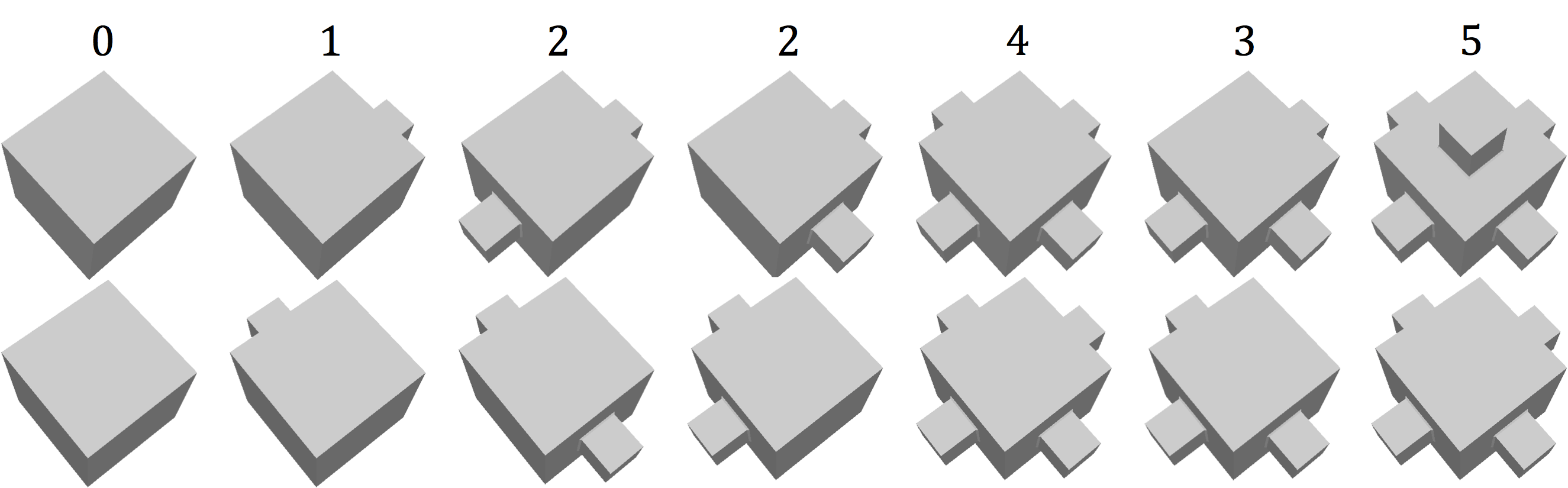}
\caption{}
\end{subfigure}
\begin{subfigure}[b]{0.6\linewidth}
\centering
\includegraphics[width=\linewidth]{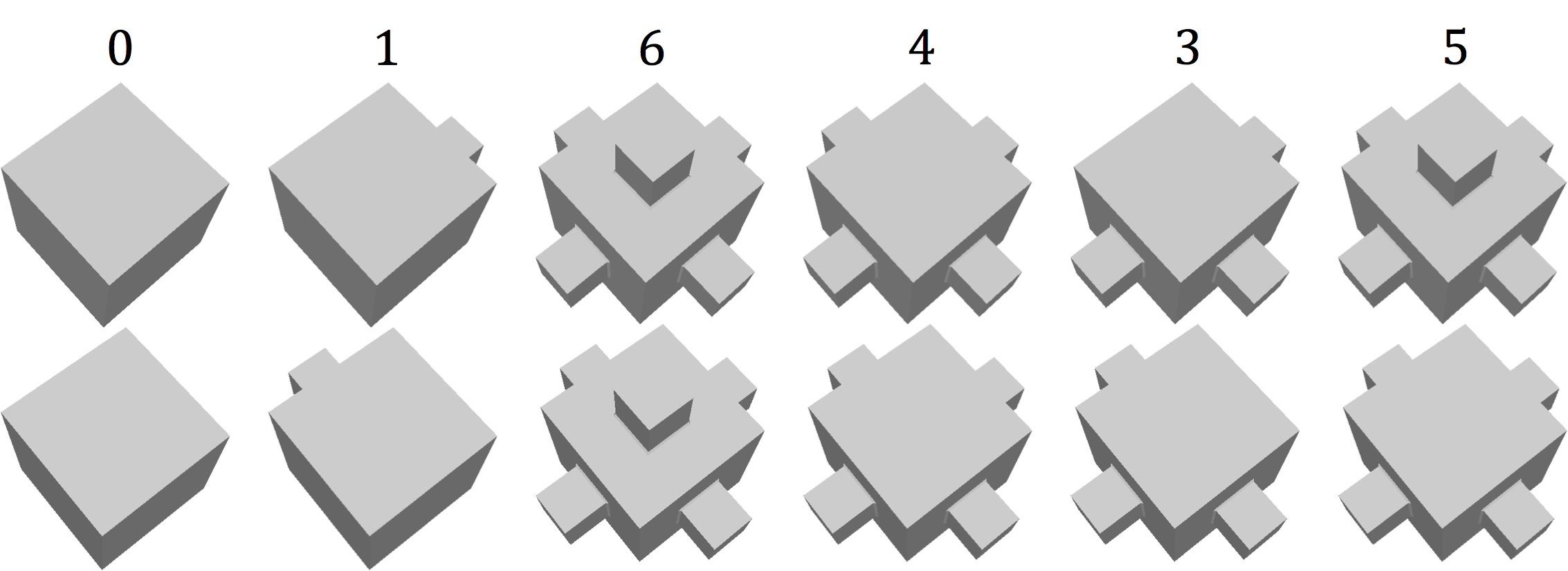}
\caption{}
\end{subfigure}
\begin{subfigure}[b]{0.8\linewidth}
\centering
\includegraphics[width=\linewidth]{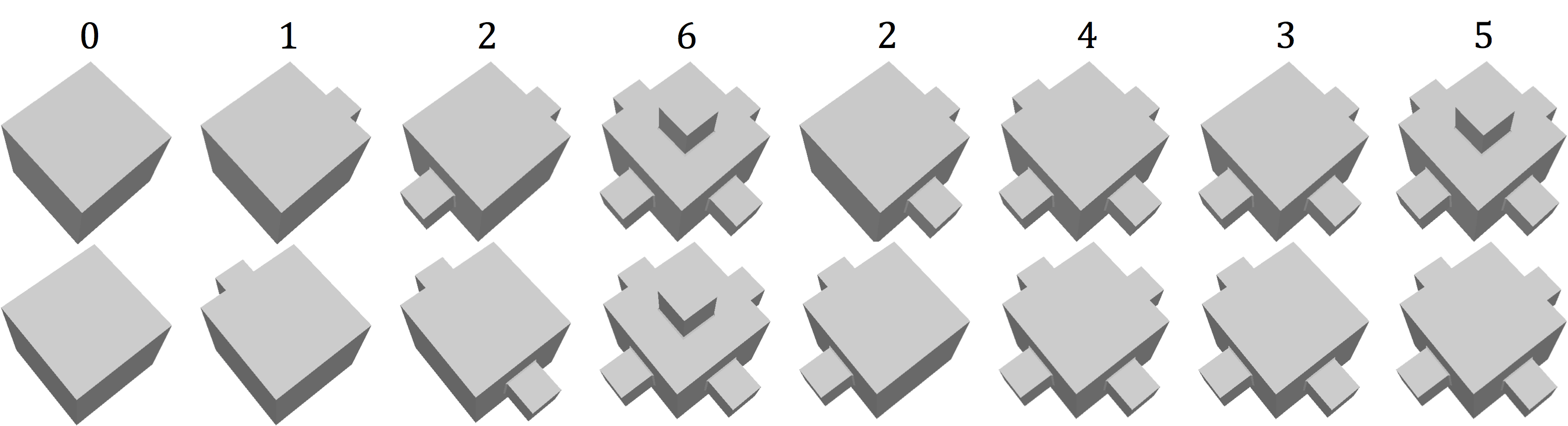}
\caption{}
\end{subfigure}
\caption{(a) and (b) Two different subsets of the shapes where the presented order remains the same for all four entropy measurements obtained using two different selections for $\rho_i$ and $\delta_k$. (c) The entire set of shapes in the order of mean value of the four entropy measures.}
\label{fig_cubes1}
\end{figure*}

\begin{figure}[t]
\centering
\includegraphics[width=0.98\linewidth]{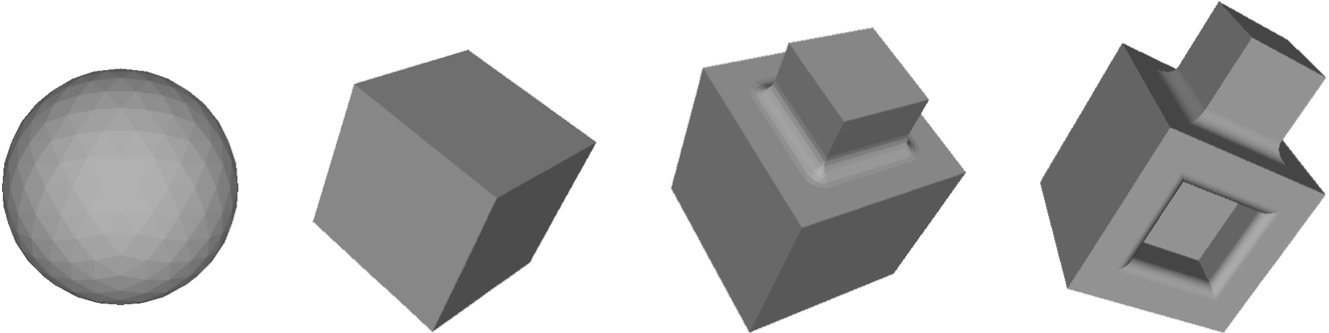}
\caption{Sample shapes whose presented order remains the same for all four entropy measures.}
\label{fig_mech} 
\end{figure}

In Fig.~\ref{fig_cubes1} (a) and (b), we present two different subsets of the first set of shapes where the presented order remains the same for all four entropy measurements. In Fig.~\ref{fig_cubes1} (c), we show the entire set ordered according to mean value of the entropy measures. We illustrate the shapes using their views from two different viewpoints in order to better distinguish them. Also, the number of cubic attachments contained in each shape is given above the corresponding views.

In Fig.~\ref{fig_cubes1} (a) and (b), we observe that our shape congruity measures tend to increase while new parts are added to a base cube in the form of smaller identical cubes as it means more deviation from a sphere. However, when the number of cubic attachments is four or six, our measures decrease due to the attained harmony. Notice that arrangement of the parts is critical for the shape congruity. For example, considering the pair of shapes with two cubic attachments, our entropy measures are smaller for the one whose attachments are facing each other. 
In Fig.~\ref{fig_vfield}, this is further illustrated using two 2D shapes that are slices extracted from the previous shape pair.
As reflected by the iso-level curves of the corresponding fields $v_{\rho_1}$ and $v_{\rho_2}$, due to arrangement of the shape parts, regions around the four corners of the larger square are identical for the first shape whereas only two of those regions are identical for the second shape, and hence, the first shape is more congruent than the second one.

We consider using mean value of the entropy measures to obtain an ordering of the entire set of the shapes which is presented in Fig.~\ref{fig_cubes1} (c). With its highly repeating structure, the shape with six cubic attachments is placed between the pair of shapes each with two cubic attachments. 

In Fig.~\ref{fig_mech}, we show the second set of shapes whose presented order remains the same for all four entropy measures. As expected, the entropy is lower for a sphere and gets higher as deformation from the prototype increases. Notice that combining a cube with a smaller cube and replacing one of its faces with a concave surface yield differing shape parts and increase deviation of shape from the prototype.

\section{Conclusion}

We presented a unified measure for congruity of shape parts and deviation of shape from a sphere i.e. our choice of the prototype. Our measure is a parameterized collection of entropy values corresponding to multiple systematic sampling of the values of a family of real-valued fields forming the shape scale space. Our entropy measure depends on two parameters which are related to smoothness of the field and common distance of the sample points to the bounding shape surface. Our measure attains its lowest value for a volume enclosed by a sphere and deformations from the prototype cause an increase in the measure. However, as deformations create congruent parts, our measure decreases due to the attained harmony.

\section*{Acknowledgment}

The work is funded by Turkish National Science Foundation TUBITAK under grant 112E208.



\bibliographystyle{IEEEtran}
%
%
%

\end{document}